\documentclass[letterpaper,10pt,conference]{ieeeconf}
\IEEEoverridecommandlockouts
\usepackage[T1]{fontenc}
\usepackage{times}
\usepackage{amsmath,amssymb}
\usepackage{graphicx,booktabs,array,tabularx}
\usepackage{placeins,stfloats}
\usepackage{xurl}
\usepackage[draft,bookmarks=false]{hyperref}
\title{\LARGE\bfseries OpenFlyScan: A Quality-Guided Aerial Reconstruction System for Consumer Drones}
\author{Zhongrui You$^{1,2}$, Zhen Li$^{2,3}$, Junli Liu$^{2,4}$, Zhigang Wang$^{2,*}$, and Bin Zhao$^{2,4}$%
\thanks{$^{1}$Beihang University, Beijing, China.}%
\thanks{$^{2}$Shanghai Artificial Intelligence Laboratory, Shanghai, China.}%
\thanks{$^{3}$Shanghai Jiao Tong University, Shanghai, China.}%
\thanks{$^{4}$Northwestern Polytechnical University, Xi'an, China.}%
\thanks{*Corresponding author: Zhigang Wang.}%
}

\begin{document}
\bstctlcite{IEEEreferencecontrol}
\maketitle
\thispagestyle{empty}
\pagestyle{empty}

\begin{abstract}
3D Gaussian Splatting (3DGS) provides high-fidelity scenes for large-scale embodied simulation, but constructing large-scale urban assets remains constrained by expensive equipment and delayed quality feedback. Preset surveys can leave complex surfaces insufficiently observed, with defects discovered only after reconstruction, requiring return visits and repeated processing. We present OpenFlyScan, a quality-guided aerial reconstruction system for consumer drones that integrates a GS quality model, a reacquisition planner, and a custom-designed mobile app. The model learns from GS rendering errors to predict regional reconstruction quality. Based on these predictions, the planner then generates complementary reacquisition strips to be executed through the app, which also supports automated oblique surveys and data transfer without additional hardware on board. Across real aerial scenes, the model effectively identifies regions that are likely to be poorly reconstructed. In the Expo West field experiment, targeted reacquisition improves PSNR at additional views by 10.95~dB. With consumer drones, OpenFlyScan integrates capture, targeted reacquisition, and reconstruction to support rapid, low-cost urban asset creation.
Code and models will be made publicly available at \url{https://openflyscan.github.io/}.
\end{abstract}

\section{Introduction}

3D Gaussian Splatting (3DGS) reconstructs real scenes from multi-view images and efficiently synthesizes realistic novel views \cite{kerbl2023gaussians}. It provides visual environments for robot simulation \cite{quach2025gsflight} and aerial vision-language navigation (VLN) \cite{gao2026openfly}. Large-area navigation and multi-agent tasks motivate extending these environments to block-scale scenes covering building facades and the ground. However, constructing these large-scale scenes remains costly in equipment and labor.

Obtaining sufficient spatial coverage and multi-view observations requires coordinated flight planning and camera settings for multidirectional oblique surveys. Such automation is primarily available on enterprise surveying platforms from manufacturers such as DJI \cite{li2026skylume,praschl2026radiometric}, whose higher purchase costs and heavier equipment burden field deployment. Consumer drones are lighter and widely available, but need corresponding planning and execution capabilities for large-scale, multidirectional capture.

Moreover, covering the survey area with preset oblique flight strips does not ensure sufficient observations of complex surfaces for high-quality 3DGS reconstruction \cite{mostegel2016uavquality}. High-rise facades and irregular structures can still suffer from occlusion, large incidence angles, and limited viewing directions, while operators cannot readily judge the need for reacquisition from transmitted images alone. Quality problems often become apparent only after full reconstruction, leading to return visits and repeated processing that increase time and labor costs \cite{lou2025ontheflyfeedback}.

Shortening this feedback cycle requires identifying regions worth reacquiring before full 3DGS reconstruction. Active capture methods provide geometric feedback through multi-view stereo (MVS) reconstruction confidence prediction \cite{mostegel2016uavquality} or mesh-quality assessment within incremental Structure-from-Motion (SfM), as in On-the-fly Feedback SfM \cite{lou2025ontheflyfeedback} (hereafter On-the-fly), but leave two issues unresolved. First, missing geometry can result from insufficient observations or failed feature matching on weakly textured surfaces and across large viewpoint changes \cite{he2024detectorfree,schoenberger2016sfm}; geometry alone cannot distinguish these causes. Second, even geometrically supported surfaces may render poorly because of appearance and illumination changes. Feed-forward 3D models rapidly provide geometry and related features \cite{wang2025vggt,wang2026pi3}, but these cues alone cannot fully reveal potential 3DGS reconstruction defects.

\begin{figure*}[!t]
\centering
\vspace*{6pt}
\includegraphics[width=\textwidth]{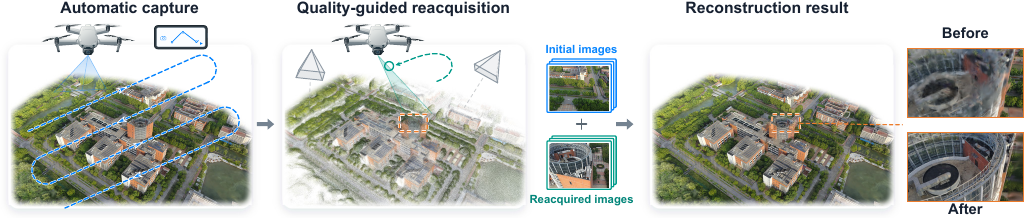}
\par\vspace{-6pt}
\caption{OpenFlyScan workflow: automated capture with consumer drones, timely regional quality feedback, targeted reacquisition, and joint 3DGS reconstruction. Routes and viewing directions are schematic.}
\label{fig:workflow}
\end{figure*}

We propose OpenFlyScan, a quality-guided aerial reconstruction system for consumer drones that integrates a GS quality model, a reacquisition planner, and a custom-designed mobile app (Fig.~\ref{fig:workflow}). Building on pretrained feed-forward geometry, our GS quality model uses a Quality Predictor to fuse multi-view image features and relative camera geometry with geometric cues, learning their relationship to regional GS reconstruction quality. Trained across a broad collection of scenes, the model provides regional quality feedback before target-scene reconstruction. Regions predicted to reconstruct poorly are selected as reacquisition targets, for which the planner arranges complementary views into continuous flight strips. Additional images are combined with the initial-survey images for reconstruction. To make this feedback and reacquisition workflow practical on consumer drones, our mobile app coordinates automated oblique surveys, data transfer, and reacquisition without additional onboard computing or communication hardware, lowering the equipment barrier to automated capture.

Across UAVFF3D-Real, Expo West, and Dishui Lake aerial surveys, our model identifies regions with high GS reconstruction error more accurately than On-the-fly. In real-world deployment, our system uses DJI Mini-series consumer drones for data capture and targeted reacquisition, completing high-quality 3DGS reconstruction of an approximately 1~km$^2$ area within 10 hours. Taking Expo West as an example, targeted reacquisition improves PSNR at additional views by 10.95~dB while improving initially weak regions.

The main contributions of this paper are:

\begin{itemize}
\item \textbf{A lightweight capture and feedback system for consumer drones.} It enables automated oblique surveys, on-site feedback, and targeted reacquisition without additional onboard hardware.
\item \textbf{A cross-modal regional quality prediction model for 3DGS reconstruction.} It combines image features, feed-forward geometry, and relative camera geometry to identify low-quality regions before target-scene 3DGS training.
\item \textbf{A quality-guided reacquisition and reconstruction workflow.} It converts quality predictions into executable flight strips, with simulation and field experiments demonstrating improvements in poorly reconstructed regions through joint reconstruction.
\end{itemize}

\section{Related Work}

\subsection{Large-Scale Urban Reconstruction and Rendering}

3DGS enables high-quality novel-view synthesis and real-time rendering \cite{kerbl2023gaussians}. Hierarchical 3DGS \cite{kerbl2024hierarchical} and CityGaussianV2 \cite{liu2025citygaussianv2} extend Gaussian reconstruction to large scenes through hierarchical representations and geometry-aware training, respectively. CityGaussianV2 reports road and facade reconstruction failures under occlusion and missing observations, while Hierarchical 3DGS identifies limited view coverage and calibration errors as sources of visual artifacts. Reliable multi-view capture therefore remains essential to constructing high-quality urban assets at scale.

\subsection{UAV Acquisition and Active View Planning}

Systematic oblique surveys coordinate flight-strip spacing, altitude, speed, and camera orientation to acquire dense multi-view observations. Enterprise platforms integrate survey planning and execution but require costly equipment: SkyLume uses a DJI M350 RTK with a C30 five-lens camera, while Praschl et al. use a DJI M30T for building reconstruction \cite{li2026skylume,praschl2026radiometric} (Table~\ref{tab:platforms}). DJI Mini drones can capture oblique views but lack an official automated oblique-survey workflow. DJI Mobile SDK exposes flight, gimbal, and camera controls, enabling automated surveys \cite{dji_msdk_software}.

\begin{table}[!ht]
\centering\footnotesize
\setlength{\tabcolsep}{4pt}
\renewcommand{\arraystretch}{1.0}
\caption{UAV acquisition platforms}
\label{tab:platforms}
\begin{tabular*}{\linewidth}{@{\extracolsep{\fill}}lrrc@{}}
\toprule
Model & \multicolumn{1}{c}{Price} & \multicolumn{1}{c}{Weight} & Official automated \\
 & \multicolumn{1}{c}{(USD)} & \multicolumn{1}{c}{(kg)} & oblique survey \\
\midrule
DJI M30T, \textit{Praschl} & 9,637 & 3.77\phantom{0} & Yes \\
DJI M350 RTK, \textit{SkyLume} & 26,470 & 6.47\phantom{0} & Yes \\
\midrule
DJI Mini 2, \textit{Ours} & 449 & <0.249 & No \\
DJI Mini 4 Pro, \textit{Ours} & 759 & <0.249 & No \\
\bottomrule
\end{tabular*}
\end{table}

Preset survey coverage does not ensure sufficient observations of complex surfaces, motivating feedback-driven view planning. SCONE and MACARONS predict additional surface coverage to guide exploration \cite{guedon2022scone,guedon2023macarons}. Mostegel et al. predict MVS reconstruction confidence from acquired images to guide subsequent observations \cite{mostegel2016uavquality}. Learning Reconstructability predicts reconstructability from proxy geometry, viewpoints, and optional images to guide urban aerial path planning \cite{liu2022reconstructability}. On-the-fly Feedback SfM combines incremental SfM, mesh-quality assessment, and predictive path planning to explore unknown areas and revisit poorly reconstructed regions \cite{lou2025ontheflyfeedback}.

Radiance-field methods also use the current model to guide acquisition. ActiveGS identifies under-reconstructed regions using Gaussian-map confidence \cite{jin2025activegs}; FisherRF estimates candidate-view information gain through Fisher information \cite{jiang2024fisherrf}. Auto3R predicts uncertainty from current GS color and depth renderings to guide scanning \cite{shen2025auto3r}. These methods require recovered geometry or a current radiance field for feedback.

\subsection{Feed-Forward Geometry and Quality Prediction}

Feed-forward models such as VGGT and $\pi^3$ enable rapid scene perception by predicting camera poses and dense geometry directly from multi-view images \cite{wang2025vggt,wang2026pi3}. UAVFF3D benchmarks these capabilities on UAV imagery and provides aerial-adapted models \cite{yang2026uavff3d}, while GeoFF3D combines coordinate anchoring and spatial chunking for large-scale aerial reconstruction \cite{yang2026geoff3d}. SwiftMap combines ORB-SLAM3 tracking and VGGT-based quality evaluation to provide on-site feedback for repeated flights \cite{xu2026swiftmap}. However, geometric confidence and coverage do not directly measure GS rendering quality under available observations.

OpenFlyScan instead predicts regional GS reconstruction quality from captured images and feed-forward geometry, before optimizing the target scene's GS representation.

\begin{figure*}[!t]
\centering
\vspace*{6pt}
\includegraphics[width=\textwidth]{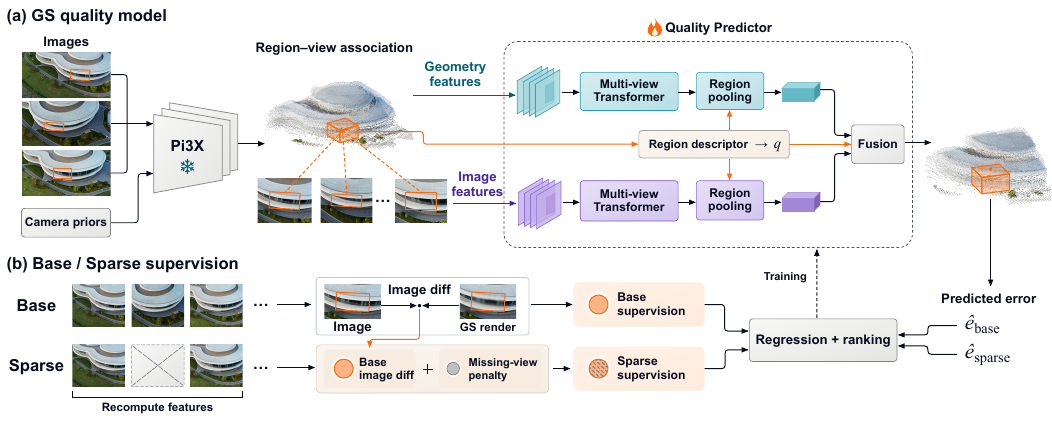}
\par\vspace{-6pt}
\caption{GS quality model: (a) regional error prediction from geometry and image features; (b) Base/Sparse supervision. Snowflake/flame: frozen/trainable parameters. Region markers and image differences are schematic.}
\label{fig:r2-head}
\end{figure*}

\section{Method}

OpenFlyScan connects initial capture with targeted reacquisition through on-site quality feedback. The mobile app manages data capture and mission execution on consumer drones, while a remote workstation predicts regional reconstruction quality before GS training and plans additional views. Initial-survey and additional images are jointly reconstructed.

\subsection{Automatic Capture and Feedback}

We develop Android and iOS applications using DJI MSDK \cite{dji_msdk_software} to enable automated multidirectional oblique surveys with consumer drones and integrate this capability with workstation-based quality assessment and reacquisition planning. We use the standard remote controller and the phone's cellular or Wi-Fi connection to communicate with the workstation, without mounting additional computing or communication hardware on the drone, as required by SwiftMap \cite{xu2026swiftmap}. The apps support 21 DJI models across the Mini, Air, Mavic, Spark, and Phantom series; our field surveys use the Mini 2 and Mini 4 Pro.

\textbf{Capture planning.} The operator specifies the survey area, flight altitude, viewing directions, and image overlap. The app projects the camera footprint onto a local horizontal plane using flight altitude, field of view, and gimbal pitch to set strip spacing and capture positions. To maintain the requested overlap, we constrain flight speed to $v\leq s_{\mathrm{photo}}/T_{\min}$, where $s_{\mathrm{photo}}$ is the capture spacing and $T_{\min}$ is the camera's minimum capture interval.

\textbf{On-site feedback.} Uploading observations during flight lets data transfer overlap with capture, reducing the wait for image export after landing. After each capture trigger, the app asynchronously reads the current video frame and uploads it with timestamped GPS and camera metadata for analysis. High-resolution images remain on the drone for final reconstruction. The workstation returns the point cloud, low-quality region markers, and proposed routes for operator review. In addition to regular survey paths, the app accepts specified camera positions, orientations, and capture actions from the workstation to execute local reacquisition.

\subsection{Regional Reconstruction Quality Prediction}

Our GS quality model consists of a pretrained Pi3X backbone and a trainable cross-modal Quality Predictor. The Quality Predictor outputs a normalized regional error $\hat e_r$ for each 3D region; larger values indicate poorer predicted quality and greater reacquisition priority. Fig.~\ref{fig:r2-head} shows the prediction pipeline and its shared Base/Sparse supervision.

\textbf{Geometry and evaluation regions.} We use UAVFF3D's UAV-adapted Pi3X model \cite{yang2026uavff3d,pi3x_software} to predict point clouds, confidence, and intermediate features from images, intrinsics, and camera pose priors derived from capture metadata. To control memory use and maintain spatial consistency, we follow GeoFF3D's spatial large-scale reconstruction framework (SLRF) \cite{yang2026geoff3d}: Pi3X processes overlapping image chunks defined by their footprints, and shared-image point correspondences provide $\mathrm{Sim}(3)$ alignment into a common frame.

Within each chunk's point cloud, a region is a small spherical neighborhood whose radius follows the local sampled-point spacing. Region--view association in Fig.~\ref{fig:r2-head}(a) uses point-to-image correspondences to find the image patches observing this neighborhood. Features at these corresponding locations describe the same local surface across views.

\textbf{Training supervision.} Fig.~\ref{fig:r2-head}(b) shows the supervision construction. Training scenes provide both captured images and completed GS reconstructions. We compare rendered images with the captured images at the original camera poses and aggregate their exposure-corrected $\log_{10}$-MSE into each 3D region. These regional errors, denoted $\overline{g}_{\mathrm{base}}$, provide Base supervision for samples with the full local image set.

To represent insufficient observation coverage, we withhold a subset of input views and rerun Pi3X on the remaining images to obtain features under reduced observations (Sparse). Region matching links each reduced-view region to its Base counterpart. Sparse supervision combines the corresponding Base rendering error with a missing-view penalty based on the fraction of supporting views withheld:
\begin{equation}
\widetilde{g}_{\mathrm{sparse}} = \overline{g}_{\mathrm{base}} + \Delta g_{\max}\left[1-\exp\left(-\frac{r_{\mathrm{drop}}}{\tau}\right)\right].
\end{equation}
Here, $r_{\mathrm{drop}}$ is the fraction of supporting views withheld. Fitting to measured GS rendering errors under reduced observations gives $\Delta g_{\max}=0.22$ and $\tau\approx0.06$. Both signals share the same training-derived normalization, mapping regional errors to $[0,1]$ while preserving order.

\textbf{Per-view features.} The geometry and image branches in Fig.~\ref{fig:r2-head}(a) first summarize observation geometry and surface appearance separately, then combine them to predict regional error. For every view observing a region, the geometry branch uses Pi3X point- and confidence-decoder features, relative camera geometry, and cross-view consistency measures to describe observation conditions and geometric agreement. The image branch uses features from the Pi3X image encoder \cite{oquab2024dinov2} and local color statistics to describe appearance. Both branches use relative camera position, viewing direction, and distance, computed from aligned predicted poses and region positions.

\textbf{Multi-view aggregation.} A region descriptor summarizes the region's position, shape, and observation statistics and is projected into query $q$. Each branch uses a one-layer, four-head Multi-view Transformer to combine information across supporting views. Region pooling uses $q$ to guide attention over valid views and produce one vector per branch. Each pooled vector also produces an initial error estimate. A two-layer fusion MLP combines these vectors, estimates, and region statistics to output the predicted regional error $\hat e_r$.

\textbf{Training.} The Pi3X backbone remains frozen. We train the shared Quality Predictor on both sample types using weighted regional regression and pairwise ranking, with auxiliary view-level supervision from measured GS errors, as shown in Fig.~\ref{fig:r2-head}(b). The overall loss is
\begin{equation}
\mathcal L=\mathcal L_{\mathrm{reg}}+\lambda_{\mathrm{rank}}\mathcal L_{\mathrm{rank}}+\lambda_{\mathrm{aux}}\mathcal L_{\mathrm{aux}}.
\end{equation}
Here, $\mathcal L_{\mathrm{reg}}$ is a Smooth L1 loss between predicted regional errors and the normalized Base/Sparse targets, while $\mathcal L_{\mathrm{rank}}$ encourages regions with larger target errors to receive higher predictions. The auxiliary loss $\mathcal L_{\mathrm{aux}}$ applies regression and ranking to view-level error predictions before regional pooling; $\lambda_{\mathrm{rank}}$ and $\lambda_{\mathrm{aux}}$ control the relative contributions of the ranking and auxiliary terms.

\subsection{Targeted Reacquisition and Joint Reconstruction}

\textbf{Reacquisition targets and candidate strips.} We merge duplicate predicted low-quality regions identified across local chunks, retain the representatives with the largest predicted errors, and select a specified number as reacquisition targets. Nearby targets that can be captured together form a capture task. We jointly plan flight strips across targets rather than selecting viewpoints independently, allowing shared observations and reducing redundant capture.

Scattered local targets use short parallel flight strips. When targets cover a larger contiguous area, the planner uses a local grid with nadir views and oblique views in four directions. Surface orientations and existing observations guide candidate strips, with capture positions determined by projected camera footprints and the requested overlap.

\textbf{Coverage and complementary observations.} We first select strips to cover the target regions and their principal viewing directions. Coverage is checked separately for each region--direction group. Strips are scored by their strongest usable view pair. A target is covered when its selected-strip score reaches 80\% of its best candidate score; each group must cover 95\% of the predicted-error weight among targets with valid candidates. Greedy selection favors coverage gained per additional image.

Coverage alone can leave support concentrated in similar directions. Candidate-view quality combines field-of-view coverage, confidence-weighted surface incidence, distance, and approximate occlusion. A strip must cover a surface sample from at least three new positions. Stereo support combines paired-view quality with an angular response over $8^\circ$--$60^\circ$. For each new view, we subtract existing nearby-direction support, estimated with a $35^\circ$ Gaussian angular kernel, from its best paired support with an existing or another new view. The largest positive difference defines the sample's added support. This favors complementary observations while retaining selected strips.

We iteratively select the feasible strip package maximizing predicted-error-weighted support gain per additional image:
\begin{equation}
A^\star=\underset{A\in\mathcal F(H)}{\arg\max}
\frac{\sum_p w_p\,g_p(A\mid H)}{n(A)}.
\end{equation}
Here, $H$ contains initial-survey observations and expected support from selected strips; $g_p(A\mid H)$ is the added directional support at surface sample $p$, and $w_p$ is its weight derived from predicted error. A candidate package $A$ includes flight strips and any intermediate rows needed for continuous scan coverage, and $n(A)$ counts all added images. The feasible set $\mathcal F(H)$ enforces image-count, path-length, estimated flight-time, and height limits. After accepting $A^\star$, we update expected support on fixed initial-survey geometry.

\textbf{Mission execution and joint reconstruction.} Selected strips are ordered to reduce transit and executed through the mobile app. Images are captured continuously along strips, not during transit. Initial camera poses remain fixed as the spatial reference. We match the additional images to feature tracks from initial-survey images and estimate their poses from 2D--3D correspondences using PnP/RANSAC \cite{schoenberger2016sfm,fischler1981ransac}. We then combine the observations, retriangulate the sparse geometry, and use OpenMVS \cite{openmvs_software} to produce a dense initialization for GS training with all images.

To reduce color-parameter storage, we retain the three base-color coefficients of degree-three spherical harmonics (SH3) \cite{kerbl2023gaussians} and encode the 45 higher-order coefficients with an eight-dimensional code and a shared $45\times8$ linear decoder. This reduces per-Gaussian color storage from 48 to 11 parameters. The codes and decoder are optimized jointly with the scene. The resulting GS scenes support UAV simulation in AirSim, built on Unreal Engine (UE) \cite{shah2017airsim}. Large-scene rendering uses hierarchical level of detail (LOD) and on-demand loading.

\section{Experiments}

We evaluate OpenFlyScan on public datasets and real aerial scenes to answer three questions: (Q1) can regional quality prediction identify poorly reconstructed GS regions more accurately than geometric quality and coverage scores; (Q2) what information do image appearance, feed-forward geometry, and relative camera geometry provide for quality prediction; and (Q3) can quality feedback guide on-site reacquisition and improve the final reconstruction?

\subsection{Experimental Setup}

We train the Quality Predictor on 17 scene groups containing 44,398 images from four datasets: UAVFF3D \cite{yang2026uavff3d}, GauU-Scene V2 \cite{xiong2024gauuscenev2}, SkyLume \cite{li2026skylume}, and UrbanScene3D \cite{lin2022urbanscene3d}. Evaluation uses three UAVFF3D-Real validation sequences and our aerial captures at Expo West and Dishui Lake, all excluded from predictor training. The Quality Predictor and its auxiliary branch are trained for 1,200 steps on 48 GPUs. The Quality Predictor has 1.20M parameters.

Field surveys use DJI Mini 2 and Mini 4 Pro drones with standard controllers and smartphones. Feed-forward geometry processing and experiments with our reconstruction backend use an RTX 4090 workstation, with eight CPU workers for cross-chunk alignment. We evaluate rendered image quality using PSNR.

\subsection{Q1: Regional Reconstruction Quality Prediction}

\textbf{Scene-wide quality ranking.}
We evaluate predicted regional errors against GS rendering errors measured after reconstruction. The reference errors are computed by comparing rendered images with captured images at the original camera poses, with exposure correction. Recall@20 measures how many of the actual highest-error 20\% of regions are recovered when selecting the predicted highest-error 20\%; Spearman's $\rho$ measures agreement across the full ranking. We compare On-the-fly \cite{lou2025ontheflyfeedback}, MACARONS \cite{guedon2023macarons}, and SwiftMap-Adapt \cite{xu2026swiftmap}. We adapt SwiftMap's publicly available modules for regional quality evaluation, forming SwiftMap-Adapt, which combines VGGT confidence with local-window pose alignment and regional aggregation. Ours (confidence-only) replaces GS quality model predictions with aggregated Pi3X geometric confidence.

As shown in Table~\ref{tab:risk-ranking}, our method achieves the highest Recall@20 and rank correlation in all three evaluation groups. Its average Recall@20 is 49.88\%, compared with 23.81\% for On-the-fly, an improvement of 26.07 percentage points. The predicted error thus prioritizes high-error regions more accurately than the geometric and coverage scores.

\begin{table}[tb]
\centering\footnotesize
\setlength{\tabcolsep}{2pt}
\renewcommand{\arraystretch}{1.0}
\vspace*{6pt}
\caption{Regional quality ranking}
\vspace{-6pt}
\label{tab:risk-ranking}
\begin{tabular*}{\linewidth}{@{\extracolsep{\fill}}llrrrr@{}}
\toprule
Method & Metric & \shortstack{UAVFF3D-\\Real (3)} & \shortstack{Expo\\West} & \shortstack{Dishui\\Lake} & All \\
\midrule
SwiftMap-Adapt & R@20$\uparrow$ & 38.82 & 23.12 & 26.45 & 30.27 \\
 & $\rho\uparrow$ & 0.389 & 0.136 & 0.226 & --- \\
\addlinespace[1pt]
On-the-fly & R@20$\uparrow$ & 26.68 & 25.36 & 20.01 & 23.81 \\
 & $\rho\uparrow$ & 0.122 & 0.096 & 0.010 & --- \\
\addlinespace[1pt]
MACARONS & R@20$\uparrow$ & 34.60 & 23.30 & 22.84 & 27.66 \\
 & $\rho\uparrow$ & 0.210 & 0.018 & 0.036 & --- \\
\midrule
Ours & R@20$\uparrow$ & 43.03 & 34.22 & 30.16 & 36.13 \\
(confidence-only) & $\rho\uparrow$ & 0.448 & 0.204 & -0.001 & --- \\
\addlinespace[1pt]
\textbf{Ours} & R@20$\uparrow$ & \textbf{51.88} & \textbf{60.51} & \textbf{40.41} & \textbf{49.88} \\
 & $\rho\uparrow$ & \textbf{0.596} & \textbf{0.680} & \textbf{0.540} & --- \\
\bottomrule
\end{tabular*}
\par\vspace{2pt}{\raggedright\footnotesize R@20 (\%). UAVFF3D-Real: three-sequence mean.\par}
\end{table}

In Dishui Lake, Ours (confidence-only) has nearly zero rank correlation with actual GS error ($\rho=-0.001$), whereas our model reaches $\rho=0.540$. Regions judged unreliable by the geometry model are not necessarily those with the poorest rendered images. In the same scene, reconstruction and alignment failures prevent On-the-fly from assigning scores to all but three evaluation regions. For evaluation, we prioritize unscored regions as potential reconstruction failures and assign them equal rank. When this tied group crosses the top-20\% cutoff, we compute Recall@20 as the expected recall under random tie-breaking.

\textbf{Identifying severely degraded surfaces.}
For target-level comparison in Dishui Lake, we evaluate Ours, Ours (confidence-only), SwiftMap-Adapt, SCONE \cite{guedon2022scone}, and MACARONS, each selecting 24 reacquisition targets. Seventeen of ours fall within the highest-error 20\% of regions, compared with 2--9 for the baselines. The median initial GS PSNR of our selected regions is 17.81 dB, compared with 21.08--22.00 dB for the baselines. Lower PSNR shows that our targets concentrate on more severely degraded surfaces.

\begin{figure}[!htbp]
\centering
\includegraphics[width=0.8\columnwidth]{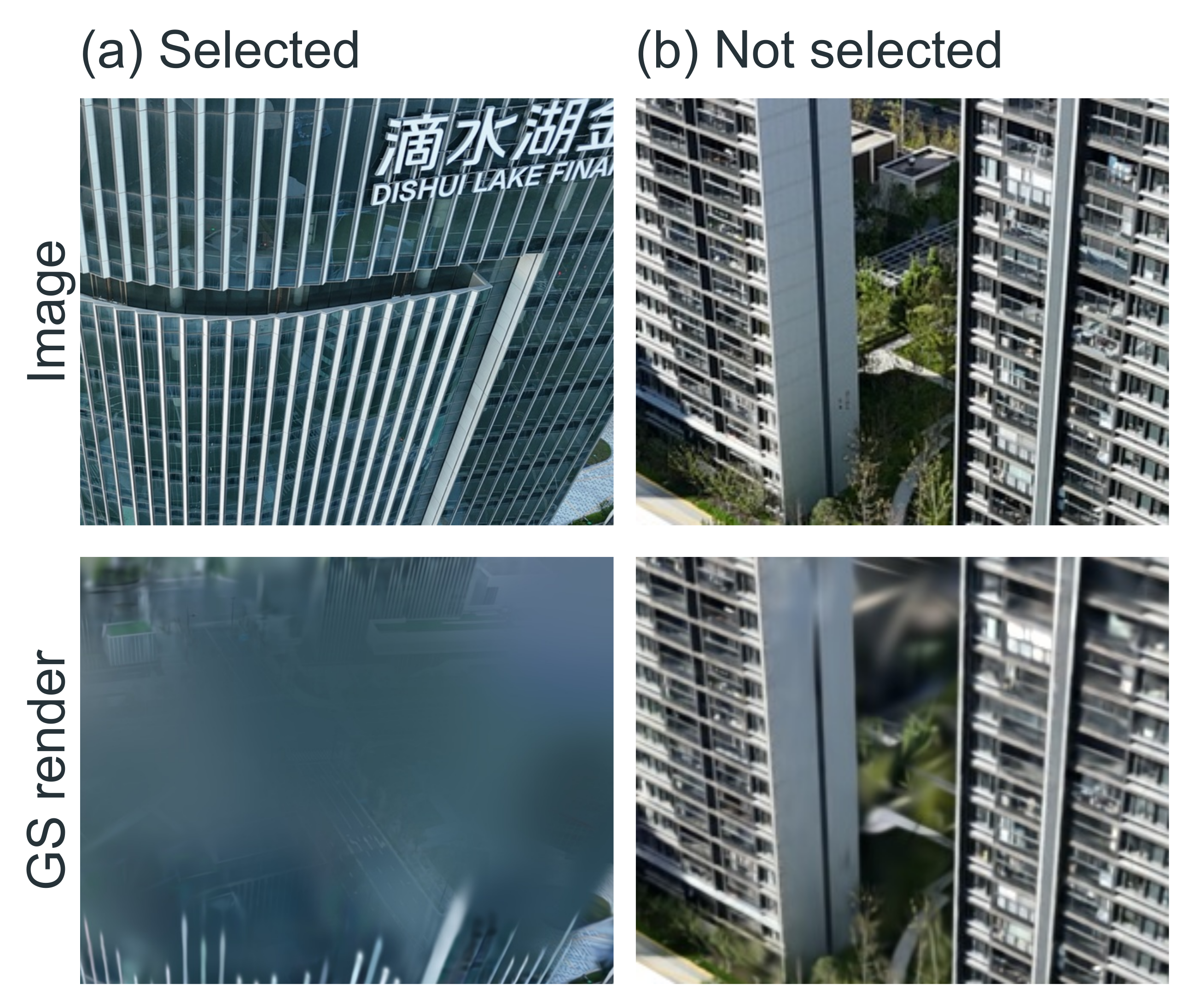}
\caption{Dishui Lake: captured images (top) and initial GS renders (bottom). (a) Ours-selected facade; (b) high-scoring facade omitted by Ours but selected by SwiftMap-Adapt.}
\label{fig:surface-comparison}
\end{figure}

Fig.~\ref{fig:surface-comparison}(a) shows a severely degraded facade selected by our method. Facade lines and signage are clear in the captured image but heavily blurred in the rendered image. Panel (b) shows a target-selection miss. Our model ranks this residential facade within the predicted highest-error 5.5\%, but spatial aggregation and budget allocation omit it from our final target set; SwiftMap-Adapt selects it. This case highlights a limitation of our target-selection strategy: spatial aggregation and budget allocation can leave severely degraded surfaces without targeted reacquisition, even when their reconstruction errors are correctly identified.

\begin{figure*}[!t]
\centering
\includegraphics[width=\textwidth]{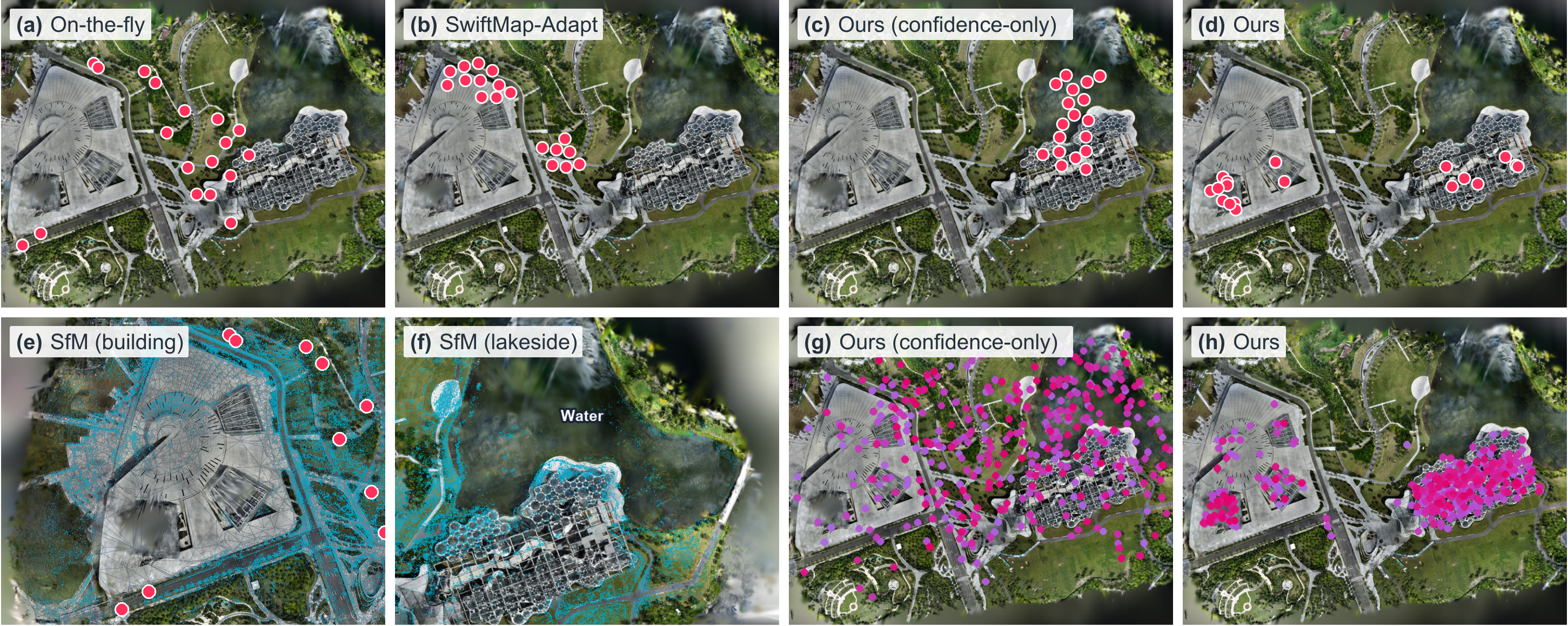}
\caption{Expo West: (a--d) 18 selected targets per method; (e,f) On-the-fly SfM support at the building and lakeside; (g,h) regional scores from Ours (confidence-only) and Ours (GS quality model), respectively. Pink: selected targets; cyan: SfM points; gray: mesh. Geometry overlays ignore occlusion.}
\label{fig:target-comparison}
\end{figure*}

\textbf{Target distribution and geometric support.}
Fig.~\ref{fig:target-comparison}(a--d) compares the reacquisition locations selected in Expo West. Our targets concentrate on the central architectural complex, while baseline targets are more dispersed across peripheral roads, vegetation, and terrain.

Fig.~\ref{fig:target-comparison}(e,f) further illustrates how missing geometric support complicates target selection. In (e), parts of the building lack SfM points, leaving geometry-based quality assessment without support on those surfaces. Yet the open water in (f) also contains point-cloud gaps; treating every unsupported region as a defect to repair would produce unnecessary reacquisition targets. Missing geometry alone therefore does not determine which surfaces warrant reacquisition.

Fig.~\ref{fig:target-comparison}(g,h) compares the regional priorities produced by the confidence-only baseline and the GS quality model. Dots mark the top 5\% of regions under each variant's ranking; for the GS quality model, violet-to-magenta indicates increasing predicted regional error. The selected regions differ, showing how learned quality prediction changes reacquisition priorities. Table~\ref{tab:risk-ranking} confirms that the GS quality model more accurately identifies high-error regions.

\subsection{Q2: Information Supporting Quality Prediction}

\textbf{Cumulative feature ablation.}
We cumulatively ablate the region-specific information in each feature group to examine what the Quality Predictor relies on. Model parameters remain fixed, and each stage retains all preceding ablations.

\begin{figure}[!htbp]
\centering
\includegraphics[width=\columnwidth]{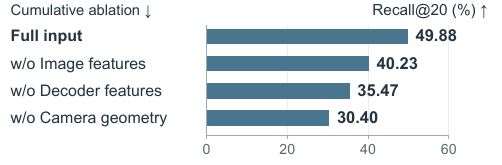}
\caption{Mean Recall@20 under cumulative feature ablation with fixed model parameters. Each w/o row additionally ablates the indicated group's region-specific information.}
\label{fig:input-sensitivity}
\end{figure}

With full input, mean Recall@20 is 49.88\%, as in Table~\ref{tab:risk-ranking}. In Fig.~\ref{fig:input-sensitivity}, recall falls to 40.23\% after ablating image features, 35.47\% after additionally ablating Pi3X point- and confidence-decoder features, and 30.40\% after also ablating relative camera geometry. The decline indicates that all three feature groups contribute to regional quality ranking.

\subsection{Q3: Reacquisition Gains and Use in UAV Simulation}

\textbf{Processing time during field operations.}
For a real outdoor survey of 1,115 images, the on-site feedback workflow from available observations to a reacquisition plan takes approximately 5 min on a single RTX 4090 workstation. Geometry processing accounts for 304.4 s, including data preparation, footprint estimation, feed-forward prediction, cross-chunk alignment, and export.

\textbf{Joint reconstruction quality and memory.}
We evaluate reconstruction quality and GPU memory on SkyLume iPark using 1,548 training images and 222 evaluation views. As shown in Table~\ref{tab:memory-quality}, under the same 10M-Gaussian cap, our backend produces fewer Gaussians than the original 3DGS with a different densification policy and contribution-based pruning, partly accounting for its lower memory use even with ordinary SH3. In our backend, the compact SH3 configuration reduces memory from 3.564 to 1.820 GiB (48.94\%) without reducing PSNR.

\begin{table}[!htbp]
\centering\footnotesize
\setlength{\tabcolsep}{3pt}
\renewcommand{\arraystretch}{1.0}
\caption{Reconstruction quality and memory}
\label{tab:memory-quality}
\begin{tabular*}{\linewidth}{@{\extracolsep{\fill}}lrrr@{}}
\toprule
Configuration & Gaussians (M) & PSNR (dB)$\uparrow$ & Memory (GiB)$\downarrow$ \\
\midrule
3DGS & 28.93 & 22.30 & 43.51 \\
3DGS (10M cap) & 9.97 & 22.75 & 15.54 \\
\midrule
Ours (SH3) & 1.32 & 22.71 & 3.56 \\
\textbf{Ours (compact)} & 1.38 & \textbf{22.94} & \textbf{1.82} \\
\bottomrule
\end{tabular*}
\par\vspace{2pt}{\raggedright\footnotesize Peak allocated memory. GPUs: A800 80GB (uncapped 3DGS); RTX 4090 (others).\par}
\end{table}

\begin{figure*}[!t]
\centering
\includegraphics[width=\textwidth]{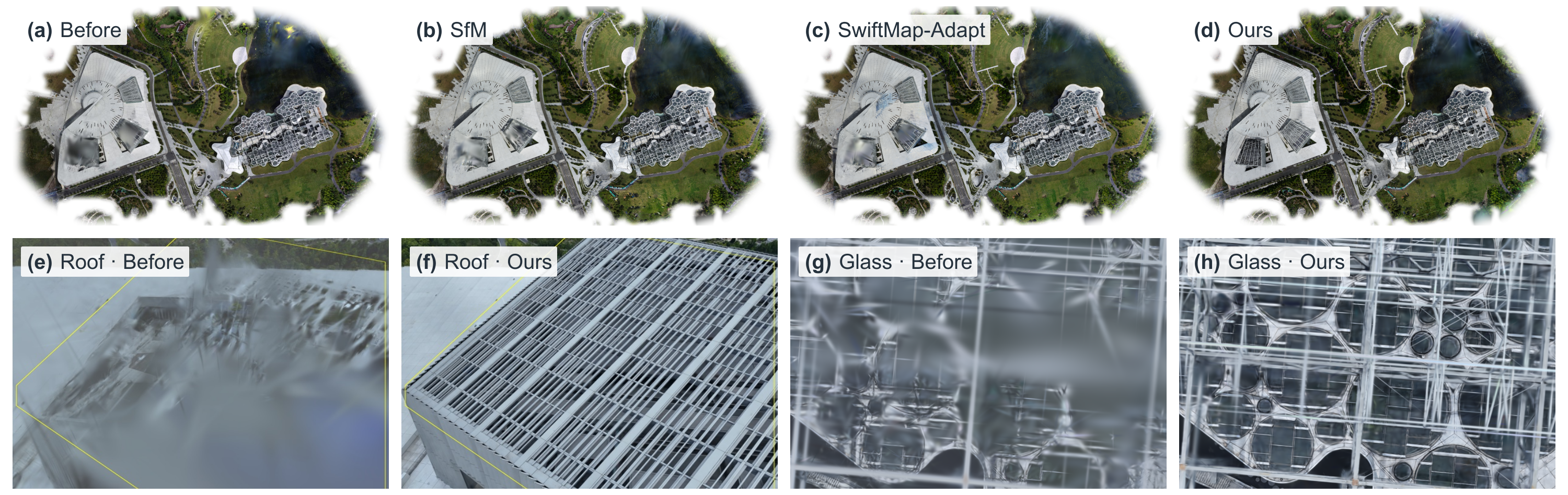}
\caption{Expo West before and after reacquisition: (a--d) scene overviews; (e--h) roof-grating and glass-roof details from our method.}
\label{fig:real-recapture}
\end{figure*}

\textbf{Reacquisition in simulation.}
We refer to the SfM-based reacquisition baselines as SfM-Q in simulation and SfM in real flight. SfM-Q combines On-the-fly quality scoring \cite{lou2025ontheflyfeedback} with the shared route generator. We compare our method with SfM-Q in NYC and CitySample. Both methods start from the same initial survey and select the same number of targets within each scene: eight in NYC and 18 in CitySample. We evaluate all initial-survey views and the worst 20\% of these views before reacquisition, keeping this subset fixed.

\begin{table}[!ht]
\centering\footnotesize
\setlength{\tabcolsep}{3pt}
\renewcommand{\arraystretch}{1.0}
\caption{Reacquisition quality (PSNR, dB)}
\label{tab:recapture-psnr}
\begin{tabular*}{\linewidth}{@{\extracolsep{\fill}}llrr@{}}
\toprule
Scene & Method & \multicolumn{1}{r}{All views$\uparrow$} & \multicolumn{1}{r}{Worst 20\%$\uparrow$} \\
\midrule
\textbf{NYC} & Before & 26.420 & 21.753 \\
 & SfM-Q & 26.605 & 22.929 \\
 & \textbf{Ours} & \textbf{26.867} & \textbf{24.088} \\
\midrule
\textbf{CitySample} & Before & 28.152 & 21.320 \\
 & SfM-Q & 28.336 & 22.513 \\
 & \textbf{Ours} & \textbf{28.604} & \textbf{23.433} \\
\bottomrule
\end{tabular*}
\end{table}

As shown in Table~\ref{tab:recapture-psnr}, our method improves PSNR on these weak views by 2.335 dB in NYC and 2.113 dB in CitySample, exceeding SfM-Q by 1.159 and 0.921 dB, respectively. More than 75\% of the initially weak views improve in both scenes, showing that the gains are not confined to a few locations.

\textbf{Planning quality and acquisition cost.}
To examine the trade-off between target-local quality and acquisition cost, we evaluate complementary-strip selection in NYC and CitySample against five-direction scanning, On-the-fly-Adapt \cite{lou2025ontheflyfeedback}, and SCONE-Adapt \cite{guedon2022scone}. On-the-fly-Adapt plans for local targets with an added two-view filter; SCONE-Adapt retains the pretrained occupancy and visibility models and coverage scoring but uses fixed geometry with greedy batch selection and route reordering. Table~\ref{tab:planner-local-efficiency} reports fixed-target crop quality and planned acquisition cost.

\begin{table}[!htbp]
\centering\footnotesize
\setlength{\tabcolsep}{2pt}
\renewcommand{\arraystretch}{1.0}
\caption{Target-local quality and acquisition cost}
\label{tab:planner-local-efficiency}
\begin{tabular*}{\linewidth}{@{\extracolsep{\fill}}llrrr@{}}
\toprule
Scene & Method & Images & Path (km) & PSNR (dB)$\uparrow$ \\
\midrule
\textbf{NYC} & Five-direction scan & 162 & 1.564 & 25.82 \\
 & On-the-fly-Adapt & 41 & 0.689 & 23.42 \\
 & SCONE-Adapt & 76 & 1.372 & 22.20 \\
 & \textbf{Ours} & 96 & 1.065 & 25.09 \\
\midrule
\textbf{CitySample} & Five-direction scan & 414 & 4.007 & 24.82 \\
 & On-the-fly-Adapt & 14 & 0.704 & 20.05 \\
 & SCONE-Adapt & 112 & 2.676 & 22.29 \\
 & \textbf{Ours} & 133 & 1.525 & 24.61 \\
\bottomrule
\end{tabular*}
\end{table}

Compared with SCONE-Adapt, our method achieves higher target-local quality with shorter planned paths in both scenes. Compared with five-direction scanning, we use 40.7\% and 67.9\% fewer images with only 0.73 and 0.21 dB lower crop PSNR. Complementary strips approach dense-scanning target-local quality with substantially fewer images.

\textbf{Real-world reacquisition.}
In the Expo West field experiment, all methods are given the same image budget for reacquisition. Each method is evaluated on initial-survey views registered in both its before and after reconstructions, together with its registered additional views. Within each paired initial-survey view set, the worst 20\% are selected by uncorrected pre-reacquisition PSNR and kept fixed. Reported PSNR uses exposure and white-balance correction.

On these fixed weak-view subsets, PSNR changes by +0.57, -0.18, and -0.45 dB for Ours, SfM, and SwiftMap-Adapt, respectively. At the additional views used in joint reconstruction, all three methods improve over their pre-reacquisition models, with gains of 10.95, 4.90, and 6.35 dB, respectively. Our method improves both newly observed areas and initially weak regions.

Fig.~\ref{fig:real-recapture} compares the full-scene reconstructions and highlights a roof grating and a glass-roof region. Our method improves reconstruction quality in these regions, producing clearer grating edges and glass-panel boundaries.

\textbf{Interactive rendering for UAV simulation.}
We evaluate rendering on a large campus scene containing approximately 72.71 million Gaussians. Drawing on Gaussian rendering approaches in NanoGS \cite{timchen_nanogs} and Spark \cite{spark_lod}, we implement hierarchical LOD selection and on-demand loading in our AirSim integration. At 1920$\times$1080 in UE 5.5 on an RTX 4090, it renders approximately 10 million active Gaussians with full SH3 colors at an average of 74.73 FPS. LumaAI 0.4.1 Niagara \cite{luma_ue041} averages 17.40 FPS with its default visible subset and static colors.

We plan to release the OpenFlyScan system, including the mobile apps, GS quality model and trained weights, the simulation framework, and GS reconstructions of all datasets used in this work.

\section{Conclusion}

OpenFlyScan connects capture with consumer drones, timely quality feedback, and targeted reacquisition in one on-site reconstruction workflow without additional onboard hardware. Experiments show that learned quality prediction improves average Recall@20 by 26.07 percentage points over On-the-fly. Targeted reacquisition improves PSNR at additional views by 10.95 dB and also improves initially weak regions. Compact color encoding reduces reconstruction memory use, while the rendering optimizations support using the scenes for UAV simulation.
Future work will improve target selection under limited acquisition budgets to reduce missed opportunities for reacquisition after quality ranking. We will explore automated collection of UAV training and evaluation data in reconstructed simulation environments.
\bibliographystyle{IEEEtran}
\bibliography{references,bibliography_control}
\end{document}